\documentclass[runningheads]{llncs}
\usepackage{graphicx}

\usepackage{tikz}
\usepackage{comment}
\usepackage{amsmath,amssymb} 
\usepackage{color}

\usepackage{graphicx}
\usepackage{comment}
\usepackage{amsmath,amssymb,amsfonts} 
\usepackage{color}
\usepackage{epsfig}
\usepackage{enumitem}
\usepackage{booktabs}
\usepackage{multirow}
\usepackage{graphicx}
\usepackage{caption}
\usepackage{mathtools}
\usepackage{url}
\usepackage{subfiles}
\usepackage{bbold}
\usepackage{microtype}
\usepackage{booktabs}
\usepackage{colortbl}
\usepackage{diagbox}
\usepackage{cite}
\usepackage[ruled,vlined]{algorithm2e}
\usepackage{soul}
\usepackage{mathrsfs}
\usepackage{paralist}

\usepackage[colorlinks,urlcolor=blue,bookmarks=false,linkcolor=blue,citecolor=blue]{hyperref}
\usepackage[capitalize]{cleveref}
\crefname{section}{Sec.}{Secs.}
\Crefname{section}{Section}{Sections}
\Crefname{table}{Table}{Tables}
\crefname{table}{Tab.}{Tabs.}

\newif\ifdrafting
\draftingtrue 
\ifdrafting
    \newcommand{\ds}[1]{{\leavevmode\color[rgb]{0,0,1}[Deqing: #1]}}
    \newcommand{\vincent}[1]{{\color{red}{\bf vincent: }#1}}
    \newcommand{\tx}[1]{{\color{cyan}{\bf Taihong: }#1}}
    
    \newcommand{\yj}[1]{{\leavevmode\color[rgb]{1,0,1}[Yaojie: #1]}}
        \newcommand{\newedit}[1]{ {\color{cyan} {#1}}}    
    
\else
    \newcommand{\ds}[1]{}
    \newcommand{\vincent}[1]{}
    \newcommand{\tx}[1]{}
    \newcommand{\yj}[1]{}
        \newcommand{\newedit}[1]{}
\fi

\makeatletter
\DeclareRobustCommand\onedot{\futurelet\@let@token\@onedot}
\def\@onedot{\ifx\@let@token.\else.\null\fi\xspace}

\def\eg{\emph{e.g}\onedot} 
\def\ie{\emph{i.e}\onedot} 
 
 \def\vs{\emph{vs}\onedot}

\def\etal{\emph{et al}\onedot}
\makeatother

\begin{document}
\pagestyle{headings}
\mainmatter
\def\ECCVSubNumber{5418}  

\title{Adaptive Transformers for Robust Few-shot Cross-domain Face Anti-spoofing} 

\titlerunning{Adaptive Transformers for Robust Few-shot Cross-domain FAS}
%

\author{Hsin-Ping Huang\inst{1} \and
Deqing Sun\inst{2} \and
Yaojie Liu\inst{2} \and
Wen-Sheng Chu\inst{2} \and
Taihong Xiao\inst{1} \and
Jinwei Yuan\inst{2} \and
Hartwig Adam\inst{2} \and
Ming-Hsuan Yang\inst{1,2,3}
}
\authorrunning{H.-P. Huang \etal}
%
\institute{$^{1}$University of California, Merced~~~$^{2}$Google Research~~~$^{3}$Yonsei University}
\maketitle

\begin{abstract}
While recent face anti-spoofing methods perform well under the intra-domain setups, an effective approach needs to account for much larger appearance variations of images acquired in complex scenes with different sensors for robust performance. 
In this paper, we present adaptive vision transformers (ViT) for robust cross-domain face anti-spoofing. 
%
%
Specifically, we adopt ViT as a backbone to exploit its strength to account for long-range dependencies among pixels. 
We further introduce the ensemble adapters module and feature-wise transformation layers in the ViT to adapt to different domains for robust performance with a few samples. 
%
%
Experiments on several benchmark datasets show that the proposed models achieve both robust and competitive performance against the state-of-the-art methods for cross-domain face anti-spoofing using a few samples. 
\end{abstract}

\section{Introduction}
\label{sec:intro}
Face biometrics is widely applied to identity authentication applications due to its security, convenience, and no-contact nature, compared to conventional methods such as passcodes and fingerprints~\cite{smartphonesecurity}.
Other than face recognition, there is an additional step needed to keep the authentication systems secure from spoof presentation attacks, which is called face anti-spoofing.
For example, printed photos, digital images, and $3$D masks can deceive mobile platforms to authenticate the attacker as the genuine user, which would cause severe security breach.
As a result, face anti-spoofing has been an important topic with studies for almost two decades. 

In early systems, face authentication is mainly applied to controlled scenarios with fixed sensors such as building access and border control. 
With controlled environments and limited variations (\eg, lighting and poses), all faces can be regarded as from one single domain. 
Numerous simple yet effective
methods~\cite{boulkenafet2015face,de2012lbp,li2016original} can be applied to determine whether spoof attacks occur or not. 
Recently, mobile applications such as unlock and payment have increased the risk of spoofing attacks. 
Face images may be acquired from wider angles, complex scenes, and different devices, which can be regarded as a set of mixed data domains.
%
In addition, an anti-spoof module may be deployed to new devices (\ie, unseen domains).
Accordingly, face anti-spoofing is required to not only handle large variations, but also well generalize or quickly adapt to unseen scenes and sensors.

Existing methods use intra-database testing and cross-database testing to evaluate the intra-domain and cross-domain face anti-spoofing performance.
%
The former one trains and evaluates models on data splits from the same database, while the latter one does from different databases.
Recent methods have already shown saturated performance on intra-database evaluations~\cite{agarwal2016face,boulkenafet2015face,atoum2017face} in well-controlled scenarios. 
In recent years, numerous methods have been proposed to tackle cross-domain face anti-spoofing~\cite{patel2016secure,liu2019deep,jia2020single}.
Although significant progress has been made, existing methods do not perform well on cross-dataset tests, \eg, on CASIA, intra-testing \vs cross-testing can be $0\%$ \vs $10\%$ on half total error rate.
%
Thus, it is of great interest to develop robust anti-spoofing methods for cross-domain scenarios. 

There are a few challenges for cross-domain face anti-spoofing applications:
\begin{compactitem}
\item \textbf{Domain gap.} The domain gap is highly correlated to the key factor of recognizing spoof: visual appearance.
Spoofing cues, such as moire pattern and color distortion, can dramatically change or disappear with different camera devices, illuminations, and image resolutions. 
For example, images from Oulu-NPU~\cite{boulkenafet2017oulu} are in 1080P resolution, while images from Idiap Replay~\cite{chingovska2012effectiveness} are only in 480P resolution. 
The sensor noise and low image quality of Idiap Replay can lead to a biased prediction as spoof from a model trained on Oulu-NPU.
\item \textbf{Limited data.} Compared to datasets for other vision tasks, \eg,  CelebA\cite{liu2015faceattributes} and FFHQ\cite{karras2019style}, commonly used datasets for face anti-spoofing (such as CASIA\cite{zhang2012face}, Idiap Replay\cite{chingovska2012effectiveness}, MSU-MFSD\cite{wen2015face}, and Oulu-NPU\cite{boulkenafet2017oulu}) are considerably smaller in scale. 
Hence, models trained with limited data can easily over-fit the training data, and thereby generalize poorly to other domains.
%
%
It is similar to training a model for downstream object recognition tasks with limited data but no ImageNet pre-trained modules.
\end{compactitem}
In this work, we address these challenges and propose a robust cross-domain model that performs as well as for intra-domain tasks. 
The proposed model learns to exploit important visual information related to face spoofing from the training data and adapt well to new domains with a few samples. 
%
%
Specifically, we introduce the vision transformer~\cite{dosovitskiy2020image} as the backbone module for cross-domain face anti-spoofing.
To facilitate cross-domain adaption with a few samples, we propose adaptive transformers by integrating ensemble adapter modules and feature-wise transformation layers.
%
%
%
Extensive experimental results show our proposed models outperform the state-of-the-art method on the widely-used benchmark datasets. 
We also provide insightful analysis on why the proposed adaptive transformer outperforms the evaluated methods. 
%
The main contributions of this work are: 
\begin{compactitem}
\item We propose adaptive transformers with ensemble adapters and feature-wise transforms for robust cross-domain face anti-spoofing with a few samples.
\item We achieve state-of-the-art cross-domain face anti-spoofing results on widely-used benchmark datasets. Our approach closes the gap between intra-database evaluation and performance in real-world applications.
\item We conduct in-depth analysis of adaptive transformers and show model explainability with insights for face anti-spoofing.

\end{compactitem}

\section{Related Work}
\label{sec:prior}

\noindent
\textbf{Face anti-spoofing.}
Early works exploit spontaneous human behaviors (\textit{e.g.}, eye blinking, head motion)~\cite{kollreider2007real,pan2007eyeblink} or predefined movements (\textit{e.g.}, head turning, expression changes)~\cite{chetty2010biometric} to address face anti-spoofing.
Due to the clear weaknesses in video replaying attacks and the inconvenience from user interaction, recent approaches evolve into modeling material properties (\ie, texture).
Several methods utilize hand-crafted features to describe spoof related patterns, \textit{e.g.}, LBP~\cite{boulkenafet2015face,de2012lbp}, HoG~\cite{komulainen2013context,yang2013face} and SIFT~\cite{patel2016secure} features, and train a live/spoof classifier using support vector machines or linear discriminant analysis. 
More recently, deep neural networks have been applied to anti-spoofing~\cite{Shao_2019_CVPR,liu2019deep,liu2020disentangling,zhang2020face,jia2020single}
and achieved state-of-the-art performance than conventional methods~\cite{yang2014learn,feng2016integration,li2016original,patel2016cross}.

As limited spoof data is available for learning classifiers or deep neural networks, auxiliary supervisory signals have been introduced to infuse the models with prior knowledge, such as facial depth map~\cite{liu2018learning}, rPPG signal~\cite{liu20163d}, reflection~\cite{yu2020face}, and face albedo~\cite{mishra2021improved}.
To improve model interpretability, feature disentanglement is proposed along with advances in generative adversarial networks~\cite{liu2020disentangling,zhang2020face}. 
Furthermore, customized deep network architectures are shown to be more effective for face anti-spoofing, \eg, tree-like structure~\cite{liu2019deep}, network architecture search~\cite{yu2020nasfas,yu2020auto}.

Most recently, a model based on a vision transformer is proposed to detect spoofing attack~\cite{george2021effectiveness}. 
Although this transformer-based method is able to detect certain spoofs, it does not perform well on challenging print and replay attacks (\eg, 5.84/15.20~\cite{george2021effectiveness} \vs 2.1/10.0~\cite{liu2019deep} EER on SiW-M dataset).
%
These results suggest large headroom to improve models in detecting low-level texture cues.
In this work, we propose an adaptive transformer model to robustly handle challenging print and replay spoof attacks across different datasets using a few-shot setting (\ie, as few as 5 samples). 

\noindent
\textbf{Domain generalization.}
Domain generalization for face anti-spoofing aims to learn a model from multiple source datasets, and the model should generalize to the unseen target dataset.
Several approaches~\cite{Shao_2019_CVPR,jia2020single,Saha_2020_CVPR_Workshops, suppressing2021} based on adversarial training and triplet loss have been developed to learn a shared feature space for multiple source domains that can generalize to the target domain. 
On the other hand, meta-learning formulations \cite{Shao_2020_AAAI,Chen2021,wang2021} are exploited to simulate the domain shift at training time to learn a representative feature space. 
Furthermore, disentangled representation for spoof and ID features \cite{padgan} and sample re-weighting \cite{Liu2021DualRD} improve generalization for face anti-spoofing.
In contrast, we tackle a real-world anti-spoofing problem when only a few images are available from target datasets.
In this work, we propose an effective cross-domain few-shot framework based on an adaptive transformer that achieves state-of-the-art performance. 

\noindent
\textbf{Few-shot learning.}
Few-shot learning methods~\cite{Hariharan_2017_ICCV,Wang_2018_CVPR,antoniou2018data,NIPS2017_cb8da676,sung2018learning,NIPS2016_90e13578,pmlr-v70-finn17a,Sachin2017} aim to adapt models to novel classes from a few samples from each class 
(assuming the classes used for training are disjoint with the novel classes seen at test time).
Cross-domain few-shot learning~\cite{Zhang_2021_ICCV,guo2020broader,crossdomainfewshot} further addresses the problem when the novel classes are sampled from a different domain with different data distribution.
In contrast, few-shot supervised domain adaptation aims to adapt models to new domains with the assistance of a few examples~\cite{NIPS2017_21c5bba1,pmlr-v119-teshima20a,Xu_2019_CVPR,motiian2017CCSA}. 
%
Anti-spoofing methods based on few-shot and zero-shot learning \cite{qin2020learning,liu2019deep} are proposed to detect multiple spoof attacks. 
The SASA method \cite{sasa} studies a similar cross-domain problem by using a few target samples to better generalize, and the features are learned within the adversarial learning framework. 
%
As shown in~\cite{sasa}, cross-domain model performance is unstable under different protocols.
In contrast, we propose to learn features from balanced data from the source domains and a few samples from the target domain.
We also propose an adaptive transformer based on an adapter and a feature-wise transformation to improve the model stability.
%
%
%

\section{Method}
\label{sec:method}

\begin{figure*}[t]
  \centering
  \includegraphics[width=1.0\linewidth]{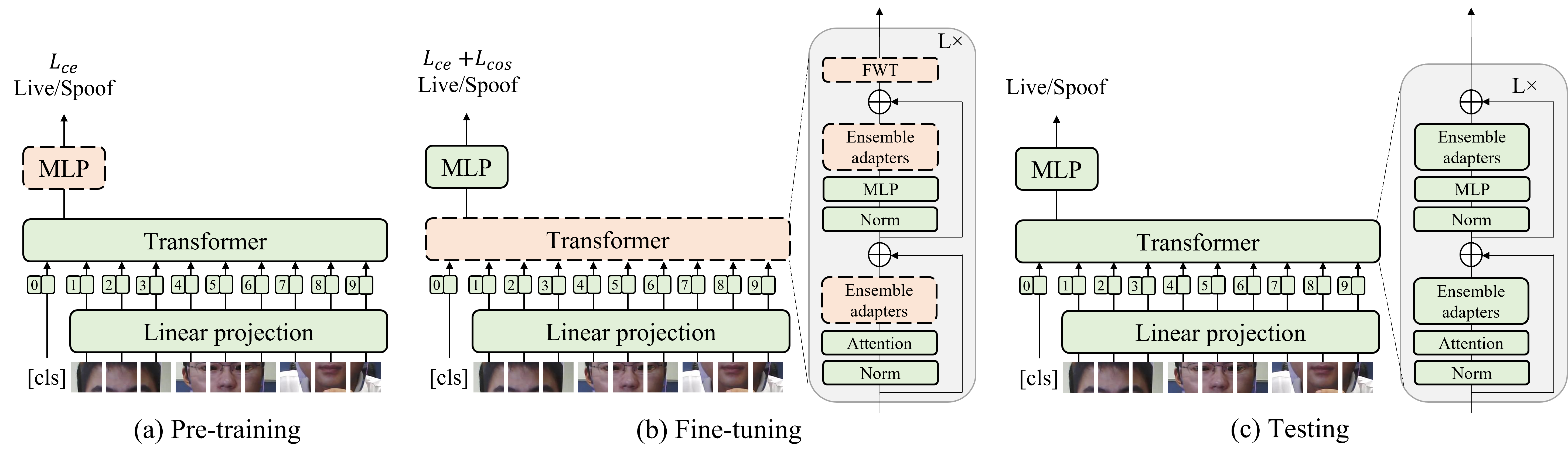}
   \caption{\textbf{Overview of our robust cross-domain face anti-spoofing framework.} The module or layer colored in wheat or green means that the weights are trainable or fixed. The transformer receives image patches and an extra learnable classification embedding [cls] as inputs, and a multi-layer perceptron (MLP) is used for live/spoof face classification. At the pre-training stage (a), the backbone is fixed and only the MLP head is trained using $L_{ce}$. At the fine-tuning stage (b), we insert two ensemble adapter modules and a feature-wise transformation (FWT) layer to each transformer block, and train all ensemble adapters and FWT layers using $L_{ce}+L_{cos}$ when other layers are fixed. During testing (c), the FWT layers are removed from the model. 
   }
   \label{fig:overview}
\end{figure*}


In this work, we assume that there exist $N$ source datasets $\mathbf{S}=\{\mathcal{S}_1, \mathcal{S}_2, \mathcal{S}_3, \ldots, \mathcal{S}_N\}$ from different domains and one target dataset $\mathcal{T}$, where each source dataset $\mathcal{S}_i$ consists of real and fake images $\mathcal{S}_i=\{X_r^{\mathcal{S}_i}, X_f^{\mathcal{S}_i}\}$.
%
The goal of few-shot cross-domain anti-spoofing is to learn a classification model that generalizes to the target domain $\mathcal{T}$ by accessing source datasets $\mathbf{S}$ as well as a few samples (\eg, 5 samples) from the target set $\mathcal{T}'=\{X_r^{\mathcal{T}'}, X_f^{\mathcal{T}'}\}\subseteq \mathcal{T}$.


 
To achieve this goal, we propose a robust framework based on the vision transformer (ViT)~\cite{dosovitskiy2020image} and the adaptive modules for few-shot cross-domain face anti-spoofing. 
%
%
The proposed approach consists of three components: vision transformer, ensemble adapters and feature-wise transformation. 
Fig.~\ref{fig:overview} shows the overall framework, and Fig.~\ref{fig:adapter} presents the adaptive modules. 
We describe each component in the following sections.

\subsection{Vision Transformer}

We adopt the vision transformers (ViT) \cite{dosovitskiy2020image} as our backbone module to tackle the face anti-spoofing problem. 
Following the standard pipeline of ViT training, we split and reshape the input image into a sequence of flattened 2D patches.  
%
%
We use ViT to obtain the image representations and a multi-layer perceptron (MLP) head to get the classification prediction, \ie, whether the input image is a live or a spoof face.
%
{
At each training iteration, we form a balanced batch by sampling the same amount of live and spoof images from $N$ source domain and a small subset of target domain $\{X_r^{\mathcal{S}_i}, X_f^{\mathcal{S}_i}, X_r^{\mathcal{T}'}, X_f^{\mathcal{T}'}\}$, where the sample size is $B$.
%
The model prediction is $\{\hat y_r^{\mathcal{S}_i}, \hat y_f^{\mathcal{S}_i}, \hat y_r^{\mathcal{T}'}, \hat y_f^{\mathcal{T}'}\}$. 
We use the cross entropy loss $L_{ce}$ to train our model, which is defined by
\begin{align}
    L_{ce} & = \frac{1}{B(N+1)}\sum_{j=1}^B
\left ( \sum_{i=1}^N(\log(\hat y_{r_j}^{\mathcal{S}_i}) + \log(1-\hat y_{f_j}^{\mathcal{S}_i})) + \log(\hat y_{r_j}^{\mathcal{T}'})+\log(1-\hat y_{f_j}^{\mathcal{T}'}) \right ).
\end{align}
}
\noindent 
%

Unlike other object classification tasks where holistic information plays an essential role, we need to detect local spoof-related cues appearing possibly all over the image for the face anti-spoofing problem. 
Empirically, it has been shown that patch-based face anti-spoofing methods~\cite{atoum2017face,8953211} improve the performance as the network extracts more discriminative local features by using patches as inputs.
However, these methods use convolutional neural networks to extract patch features and predict spoof scores for each patch independently. 
Furthermore, they use global pooling to fuse the scores for final prediction, which fails to apply global reasoning by considering the correlation among patches. 
In contrast, ViT captures the long-range dependency among different patches via the global self-attention mechanism. As a result, the local spoof-related cues can be detected independently and accumulated globally for better spoof predictions. Therefore, ViT is suitable for face anti-spoofing.
%



\subsection{Ensemble Adapters}
\label{sec:adapter}


One straightforward transfer learning strategy is to train a classifier on top of features extracted by a backbone network pre-trained on ImageNet~\cite{deng2009imagenet} using anti-spoofing data. 
However, this strategy yields poor performance on the face anti-spoofing task for two reasons. 
First, the backbone pre-trained using a generic dataset cannot adapt well to the specific anti-spoofing facial data. 
Second, features extracted from the pre-trained backbone network are high-level, thus not suitable for the face anti-spoofing task where the subtle low-level information is crucial.

Instead, one can fine-tune a classifier and the backbone on anti-spoofing data. 
Although good performance could be achieved on the source domain, the performance on the target domain becomes unstable even when the training loss approaches convergence as shown in Fig.~\ref{fig:stability} and Section \ref{sec:ablation}.
%
We attribute the instability to two factors.
1) When fine-tuning large models with few samples, the catastrophic forgetting problem usually causes training instability. 
2) The domain gap between the target samples and the source domain is large such that the target samples are close to the decision boundary and have high uncertainty.
%
An intuitive remedy is to freeze a majority of the backbone and partially fine-tune the network. 
However, the approach with fine-tuning only top layers of backbone networks does not address this issue. 
%
In the following, we propose to use ensemble adapter layers to achieve stable cross-domain performance.

\begin{figure}[t]
  \centering
  \includegraphics[width=0.35\linewidth]{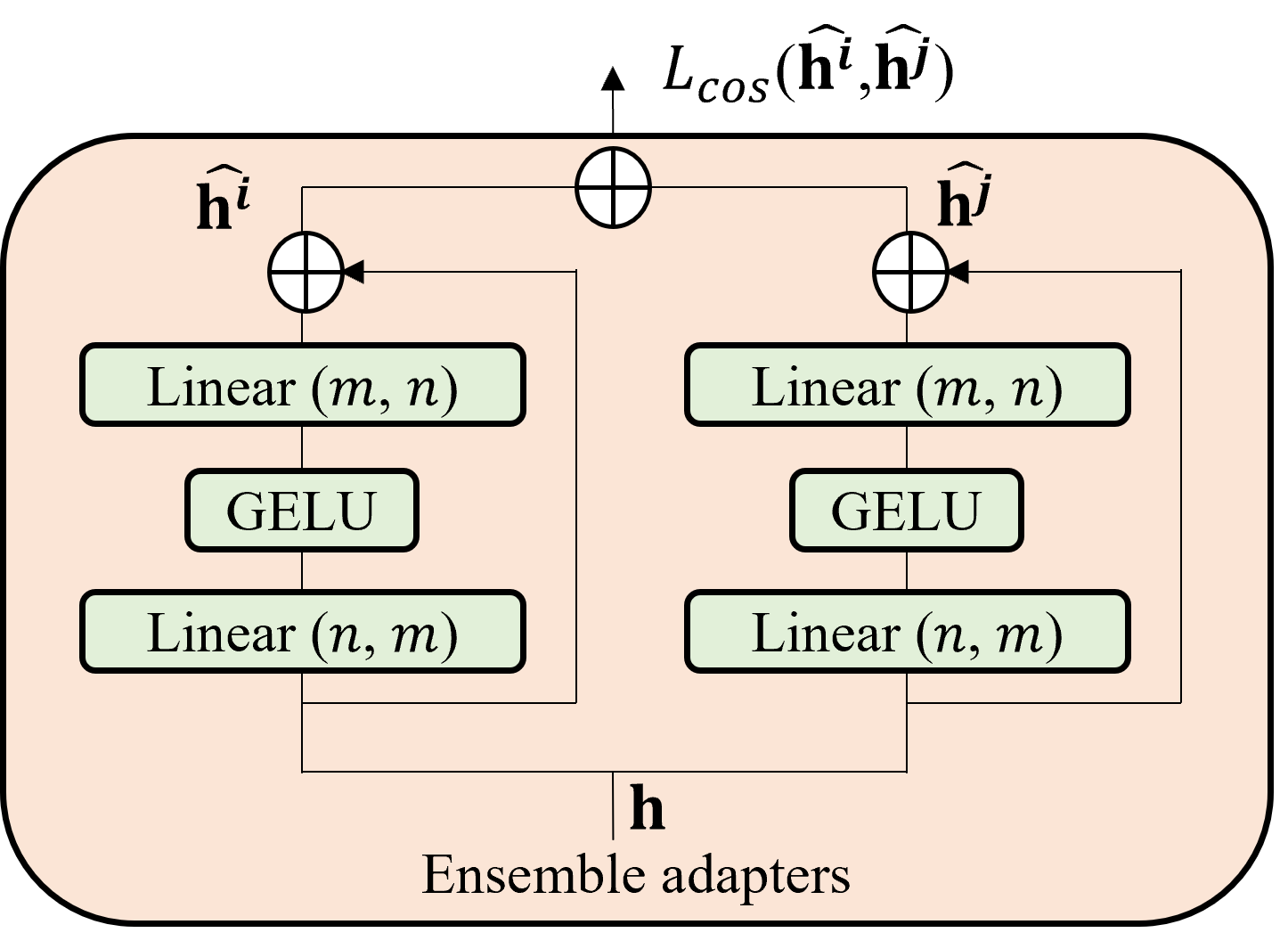}
   \caption{\textbf{Ensemble adapters.}  
 The ensemble adapters contain multiple adapters which take the same hidden representation as inputs, and the outputs of multiple adapters are aggregated. A cosine similarity loss is applied to pairs of outputs of adapters to enforce each adapter to learn diverse representations that complement with each other.
 Each adapter is a bottleneck layer with skipped connections. When the training starts, the adapter layers are close to identity layers, and they can be adapted for face anti-spoofing task or retained as the identity function.
}
   \label{fig:adapter}
\end{figure}

\Paragraph{Adaptive module.}
%
In natural language processing, 
the adapterBERT \cite{houlsby2019parameter} 
has been shown to successfully transfer the pre-trained BERT model to various downstream tasks without re-training the whole network.
In this work, we introduce the adapter layer to alleviate the instability issue. 
%
The adapter has a bottleneck architecture containing few parameters relative to the feedforward layers. 
As shown in Fig.~\ref{fig:adapter}, it first linearly projects the $n$-dimensional features into a lower dimension $m$, applies a non-linear activation function GELU, and then projects back to $n$ dimensions.
As the adapter also contains a skip-connection, it is nearly an identity function if the parameters of the projection head are initialized to near-zero.
As shown in Fig.~\ref{fig:overview}(b), two adaptive modules are inserted into each transformer block. 
During the fine-tuning stage, we fix the original transformer backbone and update the weights of adaptive modules.
%
%
As adapters contain only a few parameters ($\approx$ 3\% parameters of the original model),
they can be learned without optimization difficulties.
Thanks to the skip-connections, the adapters generate representations with less deviation from the pre-trained models and alleviate the catastrophic forgetting problem, thereby improving training stability.
%
Adapters also help adjust the feature distribution of the pre-trained transformer blocks to the face anti-spoofing data, maintaining the discriminative strength and good generalization ability of pre-trained transformer representations.

\Paragraph{Ensemble adapters and cosine similarity loss.}
In this work, we introduce the ensemble adapters module to achieve higher accuracy and minimize training instability issues. 
Instead of having two naive adapters in each transformer block, we insert two ensemble adapter modules in each block.
The ensemble adapters contain $K$ adapters in parallel.
Specifically, in each ensemble adapter module, the representation $\mathbf{h}$ is the input to $K$ adapters and the outputs of adapters $\hat {\mathbf{h}^{k}}$ are aggregated and forwarded to the next layer.
However, by simply ensembling the adapter outputs, multiple adapters learn repetitive information which does not improve the discriminability of the features and leads to limited performance improvements.
In order to learn diverse features from multiple adapters, we use a cosine similarity loss which constrains multiple outputs of adapters to be complementary.
Specifically, we minimize the cosine distance between each pair of outputs of the adapters $\hat{\mathbf{h}^{i}}$ and $\hat{\mathbf{h}^{j}}$. 
The cosine loss enforces the outputs of adapters to be dissimilar to each other and help learn diverse features.
Assume the input image has $N$ tokens and the feature dimension is $D$, we compute the cosine distance along the feature dimension and average over the number of tokens $N$. 
The cosine loss $L_{cos}$ is defined by
%
{
\begin{align}
    \hat{\mathbf{h}} & = \sum_{k=1}^K \hat{ \mathbf{h}^{k}} = \sum_{k=1}^K Adapter_k(\mathbf{h}), \\
    L_{cos} & = \sum_{1\le i,j\le K, i\neq j} \frac{1}{N} \sum^N_{n=1}\left (\frac{\hat{ \mathbf{h}_n^{i}} \cdot \hat{\mathbf{h}_n^{j}}}{\left \| \hat{ \mathbf{h}_n^{i}} \right \| \left \| \hat{ \mathbf{h}_n^{j}} \right \|}\right )^2.
\end{align}
%
As the ensemble is conducted at the bottleneck network, our ensemble module is lightweight. Adding each additional adapter requires $\approx$ 3\% additional FLOPs and parameters, which are relatively low overheads.
In practice, our model takes 15\% additional inference time compared to naive ViT.
}
\subsection{Feature-wise Transformation}

Our goal is to learn a model that generalizes well to the target domain using source datasets and a small subset of the target dataset.
Due to the distribution mismatch of the source and target domains, the model is prone to over-fitting since we only have access to limited target domain data during training. 
%
Thus, we include a feature-wise transformation (FWT) layer \cite{crossdomainfewshot} into the transformer blocks. 
%
We first sample the scaling and bias terms of affine transformations from Gaussian distributions, 
\begin{align}
    \alpha^d\sim N(0,\mathrm{softplus}(W_\alpha^d)),~~d=1, \ldots,D, \\
    \beta^d\sim N(0,\mathrm{softplus}(W_\beta^d)),~~d=1, \ldots,D,
\end{align}
\noindent where $W_\alpha^d$ and $W_\beta^d$ denote learnable sampling hyper-parameters, and $D$ denotes the channel dimension of the activation map of each transformer block.
We then compute the modulated features by applying the sampled affine transformations to intermediate features of layer $\mathbf{x}_l$ as follows: 
\begin{align}
     \hat{\mathbf{x}_l} = \mathbf{x}_l + \boldsymbol{\alpha}_l \cdot \mathbf{x}_l+\boldsymbol{\beta}_l,~~l=1,\ldots,L.
\end{align}
In practice, the same affine transformation is applied across all patch embeddings.

As shown in Fig.~\ref{fig:overview}(b), we insert one FWT layer in each transformer block.
The FWT layer is used only at training time as augmentation, and not used at test time.
The FWT layer serves as feature-level data augmentation to increase the diversity of training samples, thus dramatically reducing over-fitting and improving stability and performance. 
The FWT layer is complementary to image-level augmentation and we apply both to help model training at the same time.
%

%

\subsection{Adaptive Transformer}
The proposed adaptive transformer consists of three stages: pre-training, fine-tuning and testing, as shown in Fig.~\ref{fig:overview}.
At the pre-training stage, we fix the ViT backbone initialized with pre-trained weights from ImageNet~\cite{deng2009imagenet} and train the MLP head using the binary cross entropy loss $L_{ce}$. %
%
At the fine-tuning stage, we insert two ensemble adaptor modules and an FWT layer to each transformer block, and train all ensemble adaptors and FWT layers with all the other weights fixed until convergence using cross entropy loss and cosine loss $L_{ce}+L_{cos}$.
During the testing stage, we remove the FWT layers and keep ensemble adaptors for cross-domain classification.

\section{Experiments}
\label{sec:results}

\subsection{Experimental Setups}
\Paragraph{Datasets and protocols.} 
Two evaluation protocols are used in this work.
In \textbf{Protocol 1}, we provide evaluations on four widely-used benchmark datasets: CASIA~\cite{zhang2012face}, Idiap Replay attack~\cite{chingovska2012effectiveness}, MSU-MFSD~\cite{wen2015face}, and Oulu-NPU~\cite{boulkenafet2017oulu}.
Following the prior works, we regard each dataset as one domain and apply the leave-one-out testing protocol to evaluate the cross-domain generalization.
In \textbf{Protocol 2}, we conduct similar cross-domain evaluations on the larger-scale datasets: CASIA-SURF~\cite{Zhang_2019_CVPR,surf}, CASIA-CeFA~\cite{Liu_2021_WACV,cefa}, and WMCA~\cite{wmca}. 
Compared to datasets in \textbf{Protocol 1}, datasets in \textbf{Protocol 2} have much more subjects and richer environment variations, and thus the results can better reflect model performance. 
%
In both protocols, we include CelebA-Spoof~\cite{CelebA-Spoof} as the supplementary training data to increase the diversity of training samples to learn better representations.

\Paragraph{Implementation details.}
The input images are cropped and resized to $224\times224\times3$ and split into a patch size of $16\times16$. We use an Adam optimizer with an initial learning rate of 1e-4 and weight decay of 1e-6. The batch size is 8 for each training domain. We use ViT-Base as our backbone whose output embedding dimension is 768, and the MLP head contains two fully-connected layers whose dimensions are 512 and 2. The adapter layers have dimension $n=768$ and $m=64$. 
We set the number of ensemble adapters $K=2$ in the experiments. More experimental results of the ensemble adapters are in the supplementary materials.
We first train the MLP head for 100 iterations in the pre-training stage and then the ensemble adapters and FWT layers for 4000 iterations in the fine-tuning stage. Our method is implemented using Pytorch. 

\Paragraph{Evaluation metrics.}
We evaluate the model performance using three metrics: Half Total Error Rate (HTER), Area Under Curve (AUC), and True Positive Rate (TPR) at a fixed False Positive Rate (FPR). 
%
While HTER and AUC assess the theoretical performance, TPR at a certain FPR is adept at reflecting how well the model performs in practice. 
We use TPR@FPR=1\% as the metric which is a high usability setting.

\Paragraph{Evaluation against state-of-the-art methods.}
We evaluate our model against the state-of-the-art cross-domain face anti-spoofing SSDG~\cite{jia2020single} method under several settings:
\begin{compactitem}
    \item 0-shot SSDG\textsuperscript{\textdagger}: we do not add CelebA-Spoof, which is under the same setting as the SSDG paper~\cite{jia2020single}. 
    \item 0-shot SSDG: CelebA-Spoof is included as one of the source domains.
    \item 5-shot SSDG: CelebA-Spoof is included as one of the source domains, and the 5-shot samples are included at training time. 
\end{compactitem}
%
We include reported results of recent 0-shot methods in Table~\ref{tab:ocim} for completeness. 
These methods do not outperform our SSDG baseline model.
We are not able to evaluate these methods in the few-shot settings as the codes are not released.

\noindent \textbf{Proposed methods.}
We include these variants of our methods for evaluation:
\begin{compactitem}
    \item ViT: a ViT is used as the backbone, and the whole backbone along with the MLP layer are updated during training.
    \item ViTF: a ViT with FWT layers is used as the backbone, and the whole network along with the MLP layer are updated during training.
    \item ViTA: a ViT with naive single adapters is used as the backbone. Only the adapter layers are updated during training.
    \item ViTAF:  a ViT with naive single adapters and FWT layers is used as the backbone. Only the adapter and FWT layers are updated during training.
    \item ViTAF$^{*}$: a ViT with ensemble adapters and FWT layers is used as the backbone. The ensemble adapters and FWT layers are updated during training.
\end{compactitem}

\subsection{Cross-domain Performance}
Table \ref{tab:ocim} and Table \ref{tab:csw} show the cross-domain performance for \textbf{Protocol 1} and \textbf{Protocol 2} respectively. 
%

\Paragraph{Effectiveness of CelebA-Spoof.}
As shown in Table~\ref{tab:ocim}, the 0-shot SSDG model outperforms 0-shot SSDG\textsuperscript{\textdagger} on three out of four targets, which results in an average improvement of 0.76 AUC.
%
The improvement can be attributed to two reasons. 
First, the CelebA-Spoof dataset can increase the diversity of source domain training data.
Therefore, the distance between some source samples and the target samples is reduced, which benefits the cross-domain classification. 
Second, using more diverse training data could smooth the decision boundary and result in better model generalization ability.



\begin{table*}[!t]
\centering
\caption{Evaluation of cross-domain face anti-spoofing among CASIA (\textbf{C}), Replay (\textbf{I}), MSU-MFSD (\textbf{M}), and Oulu-NPU (\textbf{O}) databases. 
    Methods are compared at their best performance based on the evaluation process in \cite{jia2020single}.
    SSDG\textsuperscript{\textdagger} denotes the cross-domain performance reported in \cite{jia2020single} without using CelebA-Spoof as the supplementary source dataset. 
    }
\resizebox{\textwidth}{!}{
	\begin{tabular}{llccccccccccccccc}
	\multicolumn{17}{c}{} \\ \toprule
	 &\multirow{3}{*}{\bf ~~Method~} & \multicolumn{3}{c}{\textbf{OCI} $\rightarrow$ \textbf{M}}  && \multicolumn{3}{c}{\textbf{OMI} $\rightarrow$ \textbf{C}}  &&  \multicolumn{3}{c}{\textbf{OCM} $\rightarrow$ \textbf{I}} && \multicolumn{3}{c}{\textbf{ICM} $\rightarrow$ \textbf{O}} \\
	 \cmidrule{3-5} \cmidrule{7-9} \cmidrule{11-13} \cmidrule{15-17} 
	 && \multirow{2}{*}{~HTER~} & \multirow{2}{*}{~AUC~} & TPR@ && \multirow{2}{*}{~HTER~} & \multirow{2}{*}{~AUC~} & TPR@ && \multirow{2}{*}{~HTER~} & \multirow{2}{*}{~AUC~} & TPR@ && \multirow{2}{*}{~HTER~} & \multirow{2}{*}{~AUC~} & TPR@ \\
	 &&&& ~FPR=$1\%$~ &&&& ~FPR=$1\%$~ &&&& ~FPR=$1\%$~ &&&& ~FPR=$1\%$~ \\ \midrule
    &~~NAS-FAS~\cite{yu2020nasfas} & 16.85 & 90.42  & -- &&   15.21 & 92.64   & -- &&  11.63  & 96.98   & -- &&  13.16& 94.18 & --\\
    &~~DRDG~\cite{Liu2021DualRD} & 12.43 & 95.81   & -- &&  19.05& 88.79   & -- &&   15.56  &  91.79   & -- &&  15.63 & 91.75 & --\\
     0-shot &~~$D^2$AM~\cite{Chen2021} & 12.70 & 95.66  & -- &&   20.98 & 85.58   & -- &&   15.43 & 91.22  & -- &&   15.27 & 90.87 & --\\
    &~~Self-DA~\cite{wang2021} & 15.40 & 91.80 & -- && 24.50 & 84.40 & -- && 15.60 & 90.10 & -- && 23.10 & 84.30 & --\\
    &~~ANRL~\cite{anrl} & 10.83 & 96.75 & -- && 17.85 & 89.26 & -- && 16.03 & 91.04 & -- && 15.67 & 91.90 & --\\
    &~~FGHV~\cite{aaai22} & 9.17 & 96.92 & -- && 12.47 & 93.47& -- && 16.29 & 90.11 & -- && 13.58 &  93.55 &--\\
     \midrule
    &~~SSDG\textsuperscript{\textdagger}~\cite{jia2020single}
        &  7.38  &  97.17  & -- &&  10.44  &  95.94  & -- &&  11.71  &  96.59  & -- &&  15.61  &  91.54  & -- \\
    0-shot
    &~~SSDG 
        &  6.58  &  97.21  &  48.33  &&  12.91  &  93.92  &  56.43  &&  \textbf{7.01}  &  \textbf{98.28}  &  \textbf{63.85}  &&  12.47  &  94.87  &  51.55  \\
    &~~ViT
        &  \textbf{1.58}  &  \textbf{99.68}  &  \textbf{96.67}  &&   \textbf{5.70}  &  \textbf{98.91}  &  \textbf{88.57}  &&  9.25  &  97.15  &  51.54  &&  \textbf{7.47}   &  \textbf{98.42}  &  \textbf{69.30}  \\
 \midrule
    &~~SSDG 
        &  8.42  &  97.39  &  63.33  &&  12.91  &  93.59  &  60.71  &&  4.48  &  99.14  &  80.77  &&  7.81   &  97.46  &  67.61  \\
    5-shot
    &~~ViT
        &  3.42  &  98.60  &  
        {95.00}  &&  1.98  &  99.75  &  94.00  &&  {2.31}
          &  {99.75}  &  {87.69}  &&  7.34  &  97.77  &  66.90 \\
    &~~ViTA
        &  4.75  &  98.84  &  76.67  &&   5.00  &  99.13  &  82.14  &&  5.37  &  98.57  &  76.15  &&  {7.16}   &  97.97  &  {73.24}  \\
    &~~ViTAF
        &  3.42  &  99.30  &  88.33  &&   1.40  &  99.85  &  95.71  &&  3.74  &  99.34  &  85.38  &&  7.17   &  98.26  &  71.97  \\
    &~~\cellcolor{cyan}ViTAF$^{*}$
        &  \cellcolor{cyan}\textbf{2.92}  &  \cellcolor{cyan}\textbf{99.62}  &  \cellcolor{cyan}\textbf{91.66}  & \cellcolor{cyan} &   \cellcolor{cyan}\textbf{1.40}  &  \cellcolor{cyan}\textbf{99.92}  &  \cellcolor{cyan}\textbf{98.57}  & \cellcolor{cyan} &  \cellcolor{cyan}\textbf{1.64}  &  \cellcolor{cyan}\textbf{99.64} &  \cellcolor{cyan}\textbf{91.53}  & \cellcolor{cyan} &  \cellcolor{cyan}\textbf{5.39} &  \cellcolor{cyan}\textbf{98.67}  &  \cellcolor{cyan}\textbf{76.05}  \\
        \bottomrule
	\end{tabular}
\label{tab:ocim}}
\vspace{-4mm}
\end{table*}







\begin{table*}[!t]
\centering
\caption{Evaluation on cross-domain protocols among CASIA-SURF (\textbf{S}), CASIA-CeFA (\textbf{C}), and WMCA (\textbf{W}) databases. Methods are compared at their best performance based on the evaluation process in \cite{jia2020single}.}
\resizebox{0.75\textwidth}{!}{
	\begin{tabular}{llccccccccccc}
	\multicolumn{13}{c}{} \\ \toprule
	 & \multirow{3}{*}{\bf ~~Method~} & \multicolumn{3}{c}{\textbf{CS} $\rightarrow$ \textbf{W}}  && \multicolumn{3}{c}{\textbf{SW} $\rightarrow$ \textbf{C}}  &&  \multicolumn{3}{c}{\textbf{CW} $\rightarrow$ \textbf{S}} \\
	 \cmidrule{3-5} \cmidrule{7-9} \cmidrule{11-13}  
	 && \multirow{2}{*}{~HTER~} & \multirow{2}{*}{~AUC~} & TPR@ && \multirow{2}{*}{~HTER~} & \multirow{2}{*}{~AUC~} & TPR@ &&  \multirow{2}{*}{~HTER~} & \multirow{2}{*}{~AUC~} & TPR@ \\
	 &&&& ~FPR=$1\%$~ &&&& ~FPR=$1\%$~ &&&& ~FPR=$1\%$~ \\ \midrule

    \multirow{2}{*}{0-shot}
    & ~~SSDG 
        & 12.64 & 94.35 & 55.72 && 12.25 & 94.78 & \textbf{51.67} && 27.08 & 80.05 & 12.06 \\
    & ~~ViT
        & \textbf{7.98}  & \textbf{97.97} & \textbf{73.61} && \textbf{11.13} & \textbf{95.46} & 47.59 && \textbf{13.35} & \textbf{94.13} & \textbf{49.97}
    \\ \midrule
    & ~~SSDG
        & 5.08  & 99.02 & 77.49 && 6.72  & 98.11 & 74.28 && 18.88 & 88.25 & 23.42 \\
    5-shot
    & ~~ViT
        & 4.30 & 99.16 & 83.55 && 7.69 & 97.66 & 68.33 && 12.26 & {94.40} & 42.59\\
    & ~~ViTA
        & {3.77}  & 99.42 & 85.78 && {6.02}  & {98.47} & {78.29} && 15.67 & 91.86 & {51.21} \\
    & ~~ViTAF
        & 4.51  & {99.44} & {88.23} && 7.21  & 97.69 & 70.87 && {11.74} & 94.13 & 50.87 \\  
    & ~~\cellcolor{cyan}ViTAF$^{*}$
        & \cellcolor{cyan}\textbf{2.91}  & \cellcolor{cyan}\textbf{99.71} & \cellcolor{cyan}\textbf{92.65} & \cellcolor{cyan} & \cellcolor{cyan}\textbf{6.00}  & \cellcolor{cyan}\textbf{98.55} & \cellcolor{cyan}\textbf{78.56} & \cellcolor{cyan} & \cellcolor{cyan}\textbf{11.60} & \cellcolor{cyan}\textbf{95.03} & \cellcolor{cyan}\textbf{60.12} \\  
 \bottomrule
	\end{tabular}
\label{tab:csw}}
\vspace{-5mm}
\end{table*}

\Paragraph{0-shot performance.}
ViT outperforms SSDG on AUC scores for six out of total seven target domains: M (+2.5), C (+5.0), O (+3.6), W (+3.6), C (+0.7), S (+14.1), except for I (-1.1).
The result shows that ViT is a strong backbone that generalizes better to the unseen target datasets. 
A ViT backbone fine-tuned on the source datasets with only a standard cross entropy loss can achieve competitive performance upon baseline approaches that employ special domain generalization techniques such as triplet loss and adversarial training. 
We also find that adding additional triplet loss or adversarial learning does not bring performance gain to the ViT backbone.

\Paragraph{From 0-shot to 5-shot.}
5-shot SSDG improves upon 0-shot SSDG for six out of seven target domains with an average of 2.78 AUC.
Similarly, 5-shot ViT improves upon 0-shot ViT for five out of seven target domains with an average of 0.77 AUC.
%
These results demonstrate that using only 5-shot samples can effectively adapt the model to the target domain. %
Due to camera devices, illuminations, and image resolutions, a large domain gap exists between the source and target domains. In this case, a few target samples can effectively reveal some crucial characteristics and differences between live and spoof faces of the target domain, thereby facilitating the model adaptation to the target domain. 
It is worth noticing that 5-shot test samples are a relatively small subset in terms of dataset size, taking 4.7/1.6/1.2/0.3\% of the target domain data in \textbf{Protocol 1} and approximately 0.1\% of the target domain data in \textbf{Protocol 2}.

\Paragraph{5-shot performance.}
ViTA outperforms 5-shot SSDG for six out of seven target domains: M (+1.45), C (+5.54), O (+0.51), W (+0.40), C (+0.35), S (+3.60), except for I (-0.57).
Combining adapters and FWTs, the ViTAF model outperforms 5-shot SSDG for six out of seven target domains: M (+1.90), C (+6.26), O (+0.80), I (+0.20), W (+0.42), S (+5.88), except for C (-0.42).
%

\Paragraph{Comparison of ViTA and ViTAF.}
Comparing ViTA with ViTAF, FWT layers achieves consistent improvement for \textbf{Protocol 1}: M (+0.46), C (+0.72), I (+0.77), O (+0.29). 
For \textbf{Protocol 2}, ViTAF achieves improvements only for the S domain (+2.28). 
This is likely due to the smaller size of datasets in \textbf{Protocol 1}, which highlights the importance of FWT layers to generate a more diverse distribution and increase the dataset size. 

\Paragraph{Comparison of ViTAF and ViTAF$^{*}$.}
With the ensemble adapters, our full model ViTAF$^{*}$ achieves consistent improvement for all the targets over ViTAF: M (+0.32), C(+0.07), I(+0.30), O(+0.41), W(+0.27), C(+0.86), S(+0.90).
The results show that the ensemble adapters and the cosine similarity loss facilitate learning diverse features from multiple adapters which are complementary with each other.
Overall, our model achieves state-of-the-art results for all target domains, demonstrating the effectiveness of our method to adapt the model to the target domain with few samples.
We further discuss the variants of our method in the following section.

\begin{figure}[t]
  \centering
  \includegraphics[width=1.0\linewidth]{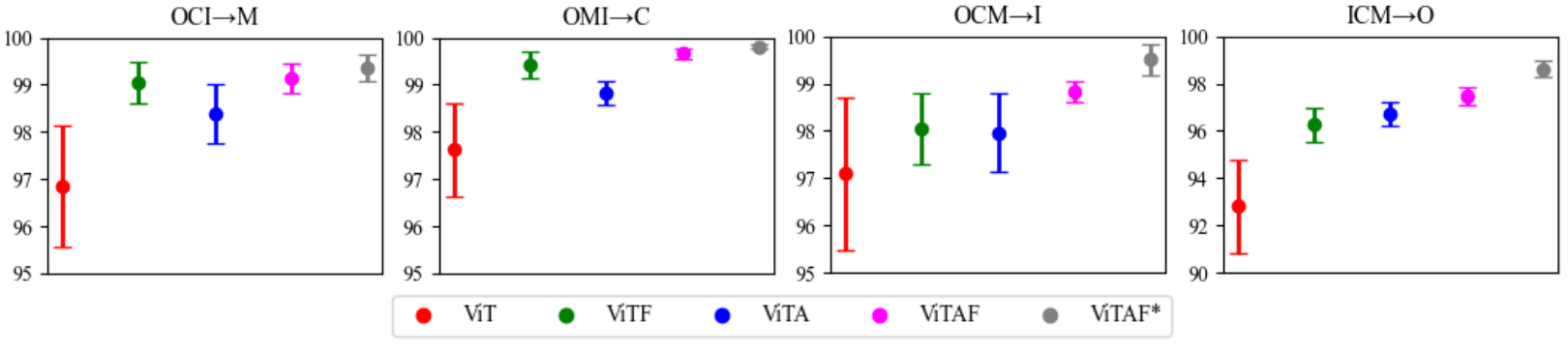}
   \caption{\textbf{Ablation study.} We analyze different components of our model including ensemble adapter layers and feature-wise transformation (FWT) layers. 
   We report the average AUC and standard deviation of the last eight checkpoints when the training is converged. The naive adapter and the FWT layer (ViTA, ViTF, ViTAF) both improve the performance. Our ensemble adapters (ViTAF$^*$) further boost the average test performance and the stability.}
   \label{fig:stability}
\end{figure}


\subsection{Ablation Study}
\label{sec:ablation}
We conduct ablation studies to analyze the contributions of each module using the 5-shot setting and \textbf{Protocol 1}. 
As discussed in Section~\ref{sec:adapter}, the performance of 5-shot ViT model may fluctuate among different checkpoints even the training loss converges. 
The best performance among all the checkpoints is reported in Table~\ref{tab:ocim} and~\ref{tab:csw}, following~\cite{jia2020single}. 
In Fig.~\ref{fig:stability}, we show the average AUC and standard deviation of the last eight checkpoints when the training process converges.
Although the \textit{best} performance of ViT is good in Table~\ref{tab:ocim}, the performance fluctuates and has lower \textit{average} performance in Fig.~\ref{fig:stability}. 
Comparing ViT with ViTF, ViTA and ViTAF, both FWT layers and adapter layers achieve better \textit{average} performance with smaller standard deviation for all targets. 
Our full model ViTAF$^{*}$ achieves the best performance with the lowest standard deviation, which validates the robust performance achieved by the ensemble adapters. 


\begin{figure*}[t]
  \centering
  \includegraphics[width=1.0\linewidth]{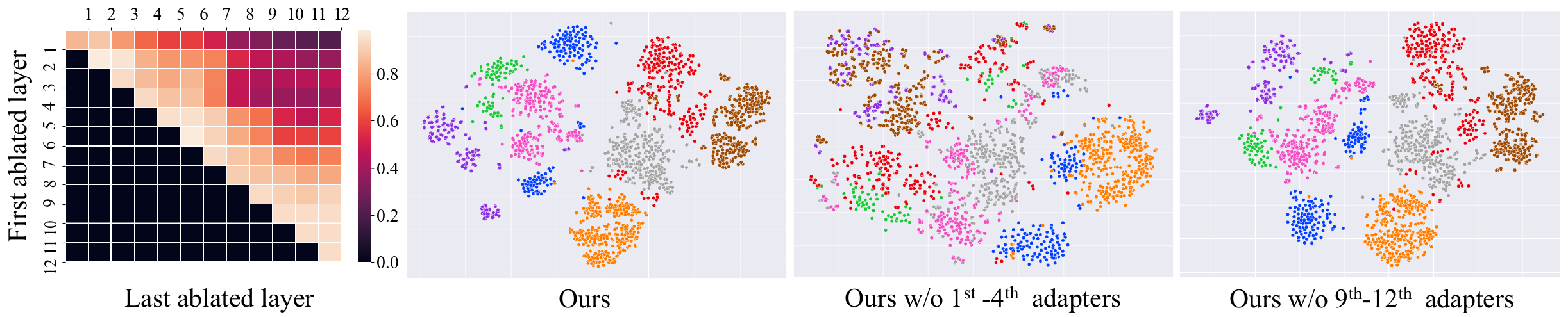}
  \caption{\textbf{Visualization of adapter.} We visualize the TPR@FPR=1\% score of ablated continuous adapter layers in the left-most sub-figure. 
  %
  For example, the value at the second row in the fifth column represents the model performance by removing the adapters from the second to fifth layers.
   %
   %
   We find that removing the adapters in the first four layers (the top left) causes a relatively severe performance drop than in the last four layers (the bottom right). 
We plot the feature distribution in the three figures to the right. A unique color represents the live or spoof sample from each domain. We observe that the data samples are less separable by removing the adapters in the first four layers than the last ones.
    }
    \label{fig:tsne}
\end{figure*}

\subsection{Visualization}

\Paragraph{Adapter.}
%
Fig.~\ref{fig:tsne} shows the TPR@FPR=1\% score of ablated continuous adapter layers. 
For example, the value at the second row in the fifth column represents the model performance by removing the adapter from the second to fifth layers.
%
%
The diagonal numbers are generally good, indicating that ablating a single adapter layer does not affect model performance significantly. 
On the other hand, removing more adapter layers causes performance drops, which can be verified by the numbers in the upper triangular region.
Moreover, the numbers at the top left are relatively smaller than those at the bottom right, which indicates that removing the adapters in the first few layers causes a more significant performance drop while removing the adapters in the last few layers does not.

We further plot the feature distribution using t-SNE~\cite{tsne}. 
We can observe that samples of all categories are well separated in our method. 
Moreover, we find that removing adapters in the first four layers leads to less separable distribution than the last four layers.
These results show that the low-level features of the transformer are more critical than high-level features for the anti-spoofing task. 


\begin{figure}[t]
  \centering \includegraphics[width=0.9\linewidth]{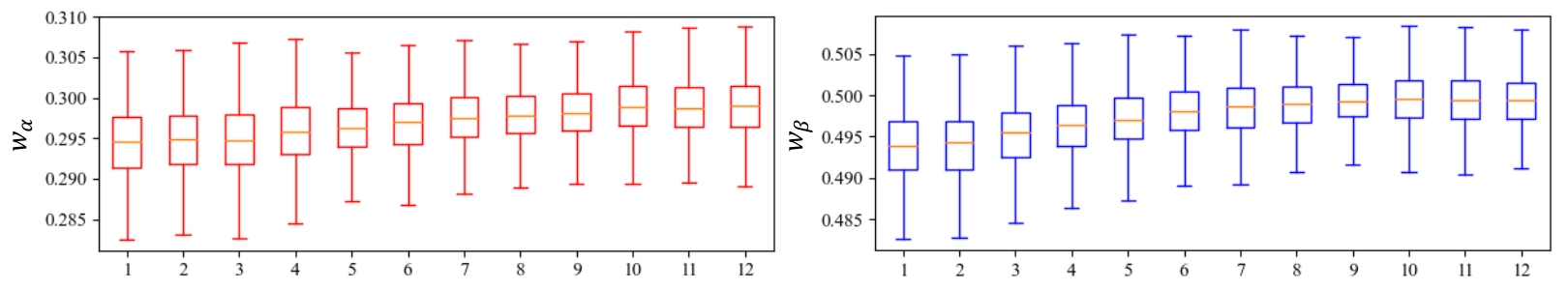}
   \caption{\textbf{Visualization of the feature-wise transformation layers.} 
   %
   We present the quartile plot of learned sampling parameters $W_\alpha^d$ and $W_\beta^d$ of all FWT layers.
   The box edges in each column mark the 25th and 75th percentiles of all $D=768$ values.
   Note that $W_\alpha^d$ and $W_\beta^d$ are initialized as 0.3 and 0.5, respectively.
   The values in shallow layers (1$^{st}$-4$^{th}$) deviate more from the initial values than those in deep layers (9$^{th}$-12$^{th}$), which suggests that the low-level features are adapted more and influence more to the target task.
   }
   \label{fig:fwt}
\end{figure}

\begin{figure}[t]
  \centering
  \includegraphics[width=0.9\linewidth]{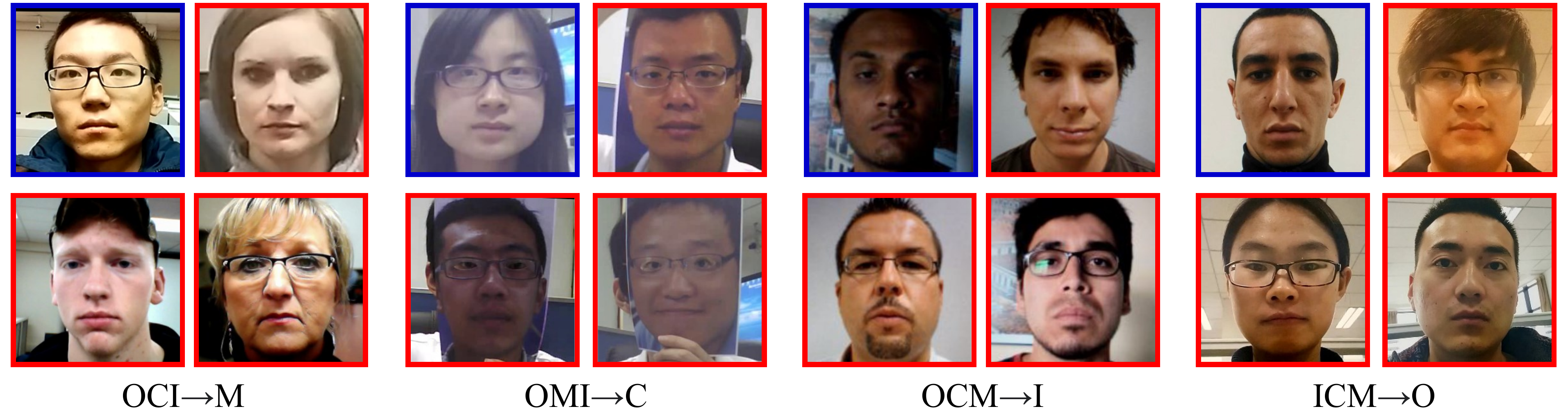}
   \caption{\textbf{Failure cases of our method.} 
   %
   The live faces misclassified as spoof faces are shown in blue boxes and the spoof faces misclassified as live faces are shown in red boxes. 
   The live faces classified as spoof faces (blue) are mostly in special light conditions or low-quality sensors, suggesting that our method still suffers from domain-specific environments. 
    The spoof faces classified as live faces (red) show that our model has difficulties detecting spoof faces with better visual quality than live faces, detecting the unseen spoof types, and detecting paper attacks with local texture differences.
   }
   \label{fig:failure}
\end{figure}

\Paragraph{Feature-wise transformation.}
Fig.~\ref{fig:fwt} shows the magnitude of the learned sampling parameters of FWT layers. We can observe that the values of both the scaling and the bias terms are closer to the initial values in deeper layers (9$^{th}$-12$^{th}$) and adjusted more in shallower layers (1$^{st}$-4$^{th}$), suggesting that the adaptation occurs mainly in the low-level features. This result also coincides with the findings in Fig.~\ref{fig:tsne}.

\Paragraph{Failure cases and limitations.}
Fig.~\ref{fig:failure} shows the common types of incorrect predictions by our methods. 
%
For the live faces classified as spoof faces (blue boxes), the results show that these faces are mostly in special light conditions or captured by low-quality sensors, \eg, strong yellow light ($\textbf{OCI} \rightarrow \textbf{M}$), sensors with weird color temperatures ($\textbf{OMI} \rightarrow \textbf{C}$), dark light ($\textbf{OCM} \rightarrow \textbf{I}$). It suggests that our method still suffers from the domain-specific light condition. Due to the extreme light condition, a live face in one domain may look like a spoof face in another domain.
For spoof faces classified as live faces (red boxes), the replay attack displaying a fixed photo is the most challenging type for $\textbf{OCM} \rightarrow \textbf{I}$. These images have even better visual quality than other live images in the same dataset, which causes confusion.
Paper attacks are more challenging for $\textbf{ICM} \rightarrow \textbf{O}$, which has the highest resolution among all datasets. 
While paper attacks show texture differences, there is no clear region of spoof cues, and thus the high-resolution spoof images in $\textbf{O}$ may look closer to live images in other datasets.
Paper attacks with real human eyes are challenging for $\textbf{OMI} \rightarrow \textbf{C}$. 
%
It is difficult to detect this specific attack that only appears in $\textbf{C}$ and does not show up in other datasets, including CelebA-Spoof.
Researchers should note that for the responsible development of this technology, it is important to consider issues of potential unfair bias and consider testing for fairness.

\begin{figure*}[t]
  \centering
  \includegraphics[width=0.9\linewidth]{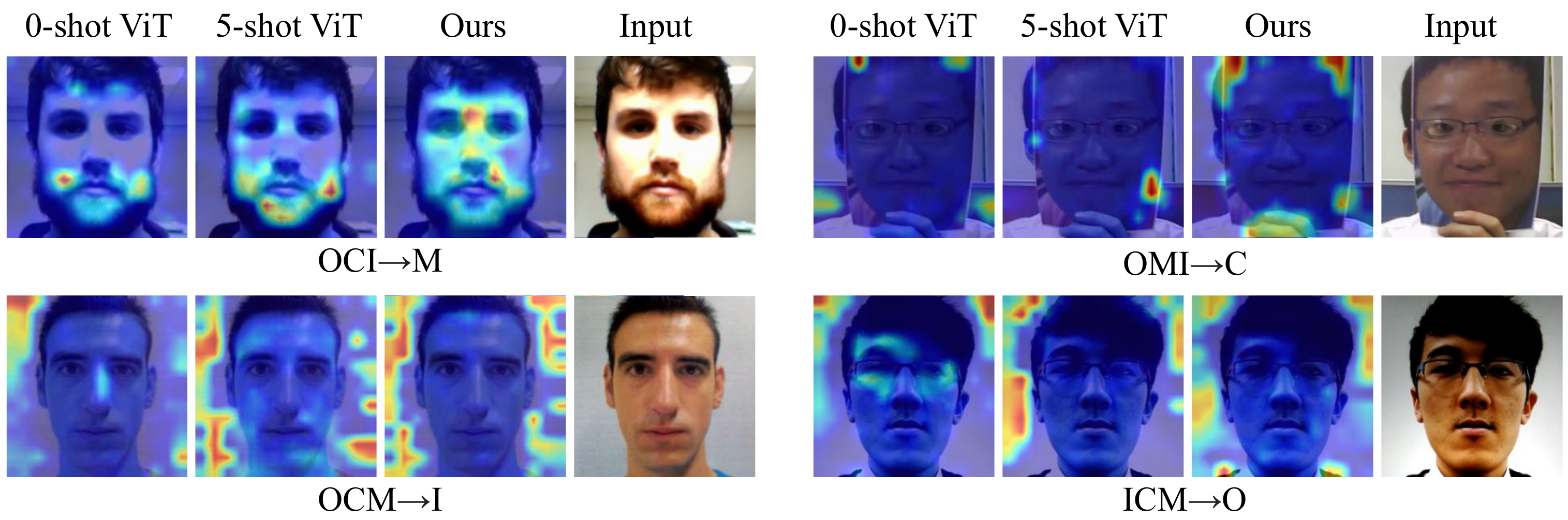}
   \caption{\textbf{Transformer attention on spoof images.} 
   We visualize the attention maps of transformers using Transformer Explainability \cite{Chefer_2021_CVPR}.
   Transformers focus on face regions with a bright reflection for MSU, hand and paper boundaries for CASIA, background paper textures for Replay, and background shadows for Oulu-NPU. 
   Our model generates more accurate and conspicuous attention maps to capture spoof cues compared with naive ViTs.
   %
   }
   \label{fig:attention}
\end{figure*}

\Paragraph{Attention maps.}
As shown in Fig.~\ref{fig:attention}, we visualize the attention maps of different transformers on spoof images using Transformer Explainability~\cite{Chefer_2021_CVPR}.
We observe that different regions are highlighted by transformers to make predictions for different spoof face domains.
%
For example, transformers make predictions mainly based on the bright reflection regions for the replay attack in $\textbf{OCI} \rightarrow \textbf{M}$. For paper attacks, transformers focus on hands and paper boundaries in $\textbf{OMI} \rightarrow \textbf{C}$, and the background paper texture in $\textbf{OCM}\rightarrow \textbf{I}$. As for the replay attack in $\textbf{ICM} \rightarrow \textbf{O}$, transformers pay more attention to the background shadows.
Moreover, our model can better capture the spoof cues compared to the naive ViT, as the attention region are more conspicuous. 
%
%

\section{Conclusion}
\label{sec:conclusion}

In this work, we study the task of cross-domain face anti-spoofing with a few samples. 
We introduce vision transformer with ensemble adapters and feature-wise transforms to adapt to new domains.
The proposed ensemble adapters significantly facilitate both stable training process and consistent model performance.
Experiments on widely-used benchmark datasets validate that our method achieves state-of-the-art performance
for few-shot cross-domain face anti-spoofing.
%
%
%



\clearpage
%
%
\bibliographystyle{splncs04}
\bibliography{egbib}

\begin{thebibliography}{10}
\providecommand{\url}[1]{\texttt{#1}}
\providecommand{\urlprefix}{URL }
\providecommand{\doi}[1]{https://doi.org/#1}

\bibitem{agarwal2016face}
Agarwal, A., Singh, R., Vatsa, M.: Face anti-spoofing using haralick features.
  In: International Conference on Biometrics Theory, Applications and Systems
  (BTAS) (2016)

\bibitem{antoniou2018data}
Antoniou, A., Storkey, A., Edwards, H.: Data augmentation generative
  adversarial networks. In: {International Conference on Learning
  Representations (ICLR)} (2018)

\bibitem{atoum2017face}
Atoum, Y., Liu, Y., Jourabloo, A., Liu, X.: Face anti-spoofing using patch and
  depth-based cnns. In: International Joint Conference on Biometrics (IJCB)
  (2017)

\bibitem{boulkenafet2015face}
Boulkenafet, Z., Komulainen, J., Hadid, A.: Face anti-spoofing based on color
  texture analysis. In: International Conference on Image Processing (ICIP)
  (2015)

\bibitem{boulkenafet2017oulu}
Boulkenafet, Z., Komulainen, J., Li, L., Feng, X., Hadid, A.: Oulu-npu: A
  mobile face presentation attack database with real-world variations. In:
  International Conference on Automatic Face \& Gesture Recognition (FG) (2017)

\bibitem{Chefer_2021_CVPR}
Chefer, H., Gur, S., Wolf, L.: Transformer interpretability beyond attention
  visualization. In: {IEEE Conference on Computer Vision and Pattern
  Recognition (CVPR)} (2021)

\bibitem{Chen2021}
Chen, Z., Yao, T., Sheng, K., Ding, S., Tai, Y., Li, J., Huang, F., Jin, X.:
  Generalizable representation learning for mixture domain face anti-spoofing.
  In: {Association for the Advancement of Artificial Intelligence (AAAI)}
  (2021)

\bibitem{chetty2010biometric}
Chetty, G.: Biometric liveness checking using multimodal fuzzy fusion. In: IEEE
  International Conference on Fuzzy Systems (FUZZ-IEEE) (2010)

\bibitem{chingovska2012effectiveness}
Chingovska, I., Anjos, A., Marcel, S.: On the effectiveness of local binary
  patterns in face anti-spoofing. In: Proceedings of the International
  Conference of Biometrics Special Interest Group (BIOSIG) (2012)

\bibitem{deng2009imagenet}
Deng, J., Dong, W., Socher, R., Li, L.J., Li, K., Fei-Fei, L.: Imagenet: A
  large-scale hierarchical image database. In: {IEEE Conference on Computer
  Vision and Pattern Recognition (CVPR)} (2009)

\bibitem{dosovitskiy2020image}
Dosovitskiy, A., Beyer, L., Kolesnikov, A., Weissenborn, D., Zhai, X.,
  Unterthiner, T., Dehghani, M., Minderer, M., Heigold, G., Gelly, S., et~al.:
  An image is worth 16x16 words: Transformers for image recognition at scale.
  In: {International Conference on Learning Representations (ICLR)} (2021)

\bibitem{feng2016integration}
Feng, L., Po, L.M., Li, Y., Xu, X., Yuan, F., Cheung, T.C.H., Cheung, K.W.:
  Integration of image quality and motion cues for face anti-spoofing: A neural
  network approach. Journal of Visual Communication and Image Representation
  \textbf{38},  451--460 (2016)

\bibitem{pmlr-v70-finn17a}
Finn, C., Abbeel, P., Levine, S.: Model-agnostic meta-learning for fast
  adaptation of deep networks. In: {International Conference on Machine
  Learning (ICML)} (2017)

\bibitem{de2012lbp}
de~Freitas~Pereira, T., Anjos, A., De~Martino, J.M., Marcel, S.: {LBP}-{TOP}
  based countermeasure against face spoofing attacks. In: {Asian Conference on
  Computer Vision (ACCV)} (2012)

\bibitem{george2021effectiveness}
George, A., Marcel, S.: On the effectiveness of vision transformers for
  zero-shot face anti-spoofing. In: International Joint Conference on
  Biometrics (IJCB) (2021)

\bibitem{wmca}
George, A., Mostaani, Z., Geissenbuhler, D., Nikisins, O., Anjos, A., Marcel,
  S.: Biometric face presentation attack detection with multi-channel
  convolutional neural network. IEEE Transactions on Information Forensics and
  Security  \textbf{15},  42--55 (2020)

\bibitem{guo2020broader}
Guo, Y., Codella, N.C., Karlinsky, L., Codella, J.V., Smith, J.R., Saenko, K.,
  Rosing, T., Feris, R.: A broader study of cross-domain few-shot learning. In:
  {European Conference on Computer Vision (ECCV)} (2020)

\bibitem{Hariharan_2017_ICCV}
Hariharan, B., Girshick, R.: Low-shot visual recognition by shrinking and
  hallucinating features. In: {IEEE International Conference on Computer Vision
  (ICCV)} (2017)

\bibitem{houlsby2019parameter}
Houlsby, N., Giurgiu, A., Jastrzebski, S., Morrone, B., De~Laroussilhe, Q.,
  Gesmundo, A., Attariyan, M., Gelly, S.: Parameter-efficient transfer learning
  for {NLP}. In: {International Conference on Machine Learning (ICML)} (2019)

\bibitem{jia2020single}
Jia, Y., Zhang, J., Shan, S., Chen, X.: Single-side domain generalization for
  face anti-spoofing. In: {IEEE Conference on Computer Vision and Pattern
  Recognition (CVPR)} (2020)

\bibitem{karras2019style}
Karras, T., Laine, S., Aila, T.: A style-based generator architecture for
  generative adversarial networks. In: {IEEE Conference on Computer Vision and
  Pattern Recognition (CVPR)} (2019)

\bibitem{suppressing2021}
Kim, T., Kim, Y.: Suppressing spoof-irrelevant factors for domain-agnostic face
  anti-spoofing. IEEE Access  (2021)

\bibitem{kollreider2007real}
Kollreider, K., Fronthaler, H., Faraj, M.I., Bigun, J.: Real-time face
  detection and motion analysis with application in “liveness” assessment.
  Transactions on Information Forensics and Security (TIFS)  \textbf{2}(3),
  548--558 (2007)

\bibitem{smartphonesecurity}
Komando, K.: Smartphone security: What’s better to use a pin, facial
  recognition, or your fingerprint? Fox News  (2019)

\bibitem{komulainen2013context}
Komulainen, J., Hadid, A., Pietik{\"a}inen, M.: Context based face
  anti-spoofing. In: International Conference on Biometrics: Theory,
  Applications and Systems (BTAS) (2013)

\bibitem{li2016original}
Li, L., Feng, X., Boulkenafet, Z., Xia, Z., Li, M., Hadid, A.: An original face
  anti-spoofing approach using partial convolutional neural network. In:
  International Conference on Image Processing Theory, Tools and Applications
  (IPTA) (2016)

\bibitem{Liu_2021_WACV}
Liu, A., Tan, Z., Wan, J., Escalera, S., Guo, G., Li, S.Z.: Casia-surf cefa: A
  benchmark for multi-modal cross-ethnicity face anti-spoofing. In: {Winter
  Conference on Applications of Computer Vision (WACV)} (2021)

\bibitem{cefa}
Liu, A., Xuan, L., Wan, J., Liang, Y., Escalera, S., Escalante, H.J., Madadi,
  M., Jin, Y., Wu, Z., Yu, X., Tan, Z., Yuan, Q., Yang, R., Zhou, B., Guo, G.,
  Li, S.: Cross‐ethnicity face anti‐spoofing recognition challenge: A
  review. IET Biometrics  \textbf{10} (2020)

\bibitem{aaai22}
Liu, S., Lu, S., Xu, H., Yang, J., Ding, S., Ma, L.: Feature generation and
  hypothesis verification for reliable face anti-spoofing. In: {Association for
  the Advancement of Artificial Intelligence (AAAI)} (2022)

\bibitem{anrl}
Liu, S., Zhang, K.Y., Yao, T., Bi, M., Ding, S., Li, J., Huang, F., Ma, L.:
  Adaptive normalized representation learning for generalizable face
  anti-spoofing. In: {ACM International Conference on Multimedia (ACM MM)}
  (2021)

\bibitem{Liu2021DualRD}
Liu, S., Zhang, K.Y., Yao, T., Sheng, K., Ding, S., Tai, Y., Li, J., Xie, Y.,
  Ma, L.: Dual reweighting domain generalization for face presentation attack
  detection. In: {International Joint Conference on Artificial Intelligence
  (IJCAI)} (2021)

\bibitem{liu20163d}
Liu, S., Yuen, P.C., Zhang, S., Zhao, G.: 3d mask face anti-spoofing with
  remote photoplethysmography. In: {European Conference on Computer Vision
  (ECCV)} (2016)

\bibitem{liu2018learning}
Liu, Y., Jourabloo, A., Liu, X.: Learning deep models for face anti-spoofing:
  Binary or auxiliary supervision. In: {IEEE Conference on Computer Vision and
  Pattern Recognition (CVPR)} (2018)

\bibitem{liu2019deep}
Liu, Y., Stehouwer, J., Jourabloo, A., Liu, X.: Deep tree learning for
  zero-shot face anti-spoofing. In: {IEEE Conference on Computer Vision and
  Pattern Recognition (CVPR)} (2019)

\bibitem{liu2020disentangling}
Liu, Y., Stehouwer, J., Liu, X.: On disentangling spoof trace for generic face
  anti-spoofing. In: {European Conference on Computer Vision (ECCV)} (2020)

\bibitem{liu2015faceattributes}
Liu, Z., Luo, P., Wang, X., Tang, X.: Deep learning face attributes in the
  wild. In: {IEEE International Conference on Computer Vision (ICCV)} (2015)

\bibitem{tsne}
van~der Maaten, L., Hinton, G.: Viualizing data using t-sne. {Journal of
  Machine Learning Research (JMLR)}  \textbf{9},  2579--2605 (2008)

\bibitem{mishra2021improved}
Mishra, S.K., Sengupta, K., Horowitz-Gelb, M., Chu, W.S., Bouaziz, S., Jacobs,
  D.: Improved detection of face presentation attacks using image
  decomposition. arXiv preprint arXiv:2103.12201  (2021)

\bibitem{NIPS2017_21c5bba1}
Motiian, S., Jones, Q., Iranmanesh, S., Doretto, G.: Few-shot adversarial
  domain adaptation. In: {Neural Information Processing Systems (NeurIPS)}
  (2017)

\bibitem{motiian2017CCSA}
Motiian, S., Piccirilli, M., Adjeroh, D.A., Doretto, G.: Unified deep
  supervised domain adaptation and generalization. In: {IEEE International
  Conference on Computer Vision (ICCV)} (2017)

\bibitem{pan2007eyeblink}
Pan, G., Sun, L., Wu, Z., Lao, S.: Eyeblink-based anti-spoofing in face
  recognition from a generic webcamera. In: {IEEE International Conference on
  Computer Vision (ICCV)} (2007)

\bibitem{patel2016cross}
Patel, K., Han, H., Jain, A.K.: Cross-database face antispoofing with robust
  feature representation. In: Chinese Conference on Biometric Recognition
  (CCBR) (2016)

\bibitem{patel2016secure}
Patel, K., Han, H., Jain, A.K.: Secure face unlock: Spoof detection on
  smartphones. Transactions on Information Forensics and Security (TIFS)
  \textbf{11}(10),  2268--2283 (2016)

\bibitem{qin2020learning}
Qin, Y., Zhao, C., Zhu, X., Wang, Z., Yu, Z., Fu, T., Zhou, F., Shi, J., Lei,
  Z.: Learning meta model for zero-and few-shot face antispoofing. In:
  {Association for the Advancement of Artificial Intelligence (AAAI)} (2020)

\bibitem{Sachin2017}
Ravi, S., Larochelle, H.: Optimization as a model for few-shot learning. In:
  {International Conference on Learning Representations (ICLR)} (2017)

\bibitem{Saha_2020_CVPR_Workshops}
Saha, S., Xu, W., Kanakis, M., Georgoulis, S., Chen, Y., Paudel, D.P.,
  Van~Gool, L.: Domain agnostic feature learning for image and video based face
  anti-spoofing. In: {IEEE Conference on Computer Vision and Pattern
  Recognition Workshops (CVPRW)} (2020)

\bibitem{Shao_2019_CVPR}
Shao, R., Lan, X., Li, J., Yuen, P.C.: Multi-adversarial discriminative deep
  domain generalization for face presentation attack detection. In: {IEEE
  Conference on Computer Vision and Pattern Recognition (CVPR)} (2019)

\bibitem{Shao_2020_AAAI}
Shao, R., Lan, X., Yuen, P.C.: Regularized fine-grained meta face
  anti-spoofing. In: {Association for the Advancement of Artificial
  Intelligence (AAAI)} (2020)

\bibitem{NIPS2017_cb8da676}
Snell, J., Swersky, K., Zemel, R.: Prototypical networks for few-shot learning.
  In: {Neural Information Processing Systems (NeurIPS)} (2017)

\bibitem{sung2018learning}
Sung, F., Yang, Y., Zhang, L., Xiang, T., Torr, P.H., Hospedales, T.M.:
  Learning to compare: Relation network for few-shot learning. In: {IEEE
  Conference on Computer Vision and Pattern Recognition (CVPR)} (2018)

\bibitem{pmlr-v119-teshima20a}
Teshima, T., Sato, I., Sugiyama, M.: Few-shot domain adaptation by causal
  mechanism transfer. In: {International Conference on Machine Learning (ICML)}
  (2020)

\bibitem{crossdomainfewshot}
Tseng, H.Y., Lee, H.Y., Huang, J.B., Yang, M.H.: Cross-domain few-shot
  classification via learned feature-wise transformation. In: {International
  Conference on Learning Representations (ICLR)} (2020)

\bibitem{NIPS2016_90e13578}
Vinyals, O., Blundell, C., Lillicrap, T., kavukcuoglu, k., Wierstra, D.:
  Matching networks for one shot learning. In: {Neural Information Processing
  Systems (NeurIPS)} (2016)

\bibitem{padgan}
Wang, G., Han, H., Shan, S., Chen, X.: Cross-domain face presentation attack
  detection via multi-domain disentangled representation learning. In: {IEEE
  Conference on Computer Vision and Pattern Recognition (CVPR)} (2020)

\bibitem{wang2021}
Wang, J., Zhang, J., Bian, Y., Cai, Y., Wang, C., Pu, S.: Self-domain
  adaptation for face anti-spoofing. In: {Association for the Advancement of
  Artificial Intelligence (AAAI)} (2021)

\bibitem{Wang_2018_CVPR}
Wang, Y.X., Girshick, R., Hebert, M., Hariharan, B.: Low-shot learning from
  imaginary data. In: {IEEE Conference on Computer Vision and Pattern
  Recognition (CVPR)} (2018)

\bibitem{wen2015face}
Wen, D., Han, H., Jain, A.K.: Face spoof detection with image distortion
  analysis. Transactions on Information Forensics and Security (TIFS)
  \textbf{10}(4),  746--761 (2015)

\bibitem{Xu_2019_CVPR}
Xu, X., Zhou, X., Venkatesan, R., Swaminathan, G., Majumder, O.: d-sne: Domain
  adaptation using stochastic neighborhood embedding. In: {IEEE Conference on
  Computer Vision and Pattern Recognition (CVPR)} (2019)

\bibitem{sasa}
Yang, B., Zhang, J., Yin, Z., Shao, J.: Few-shot domain expansion for face
  anti-spoofing. arXiv preprint arXiv:2106.14162  (2021)

\bibitem{yang2014learn}
Yang, J., Lei, Z., Li, S.Z.: Learn convolutional neural network for face
  anti-spoofing. arXiv preprint arXiv:1408.5601  (2014)

\bibitem{yang2013face}
Yang, J., Lei, Z., Liao, S., Li, S.Z.: Face liveness detection with component
  dependent descriptor. In: International Confernece on Biometrics (ICB) (2013)

\bibitem{8953211}
Yang, X., Luo, W., Bao, L., Gao, Y., Gong, D., Zheng, S., Li, Z., Liu, W.: Face
  anti-spoofing: Model matters, so does data. In: {IEEE Conference on Computer
  Vision and Pattern Recognition (CVPR)} (2019)

\bibitem{yu2020face}
Yu, Z., Li, X., Niu, X., Shi, J., Zhao, G.: Face anti-spoofing with human
  material perception. In: {European Conference on Computer Vision (ECCV)}
  (2020)

\bibitem{yu2020auto}
Yu, Z., Qin, Y., Xu, X., Zhao, C., Wang, Z., Lei, Z., Zhao, G.: Auto-fas:
  Searching lightweight networks for face anti-spoofing. In: {IEEE
  International Conference on Acoustics, Speech and SP (ICASSP)} (2020)

\bibitem{yu2020nasfas}
Yu, Z., Wan, J., Qin, Y., Li, X., Li, S., Zhao, G.: Nas-fas: Static-dynamic
  central difference network search for face anti-spoofing. {IEEE Transactions
  on Pattern Recognition and Machine Intelligence (PAMI)}  \textbf{43},
  3005--3023 (2021)

\bibitem{zhang2020face}
Zhang, K.Y., Yao, T., Zhang, J., Tai, Y., Ding, S., Li, J., Huang, F., Song,
  H., Ma, L.: Face anti-spoofing via disentangled representation learning. In:
  {European Conference on Computer Vision (ECCV)} (2020)

\bibitem{surf}
Zhang, S., Liu, A., Wan, J., Liang, Y., Guo, G., Escalera, S., Escalante, H.J.,
  Li, S.Z.: Casia-surf: A large-scale multi-modal benchmark for face
  anti-spoofing. IEEE Transactions on Biometrics, Behavior, and Identity
  Science (T-BIOM)  (2020)

\bibitem{Zhang_2019_CVPR}
Zhang, S., Wang, X., Liu, A., Zhao, C., Wan, J., Escalera, S., Shi, H., Wang,
  Z., Li, S.Z.: A dataset and benchmark for large-scale multi-modal face
  anti-spoofing. In: {IEEE Conference on Computer Vision and Pattern
  Recognition (CVPR)} (2019)

\bibitem{Zhang_2021_ICCV}
Zhang, X., Meng, D., Gouk, H., Hospedales, T.M.: Shallow bayesian meta learning
  for real-world few-shot recognition. In: {IEEE International Conference on
  Computer Vision (ICCV)} (2021)

\bibitem{CelebA-Spoof}
Zhang, Y., Yin, Z., Li, Y., Yin, G., Yan, J., Shao, J., Liu, Z.: Celeba-spoof:
  Large-scale face anti-spoofing dataset with rich annotations. In: {European
  Conference on Computer Vision (ECCV)} (2020)

\bibitem{zhang2012face}
Zhang, Z., Yan, J., Liu, S., Lei, Z., Yi, D., Li, S.Z.: A face antispoofing
  database with diverse attacks. In: International Confernece on Biometrics
  (ICB) (2012)

\end{thebibliography}
\end{document}